%% file: paper.tex
\documentclass[10pt,twocolumn,letterpaper]{article}

\usepackage[pagenumbers]{cvpr} 

\usepackage[accsupp]{axessibility}

\input{preamble}

\definecolor{cvprblue}{rgb}{0.21,0.49,0.74}
\usepackage[pagebackref,breaklinks,colorlinks,citecolor=cvprblue]{hyperref}

\usepackage[capitalize]{cleveref}
\crefname{section}{Sec.}{Secs.}
\Crefname{section}{Section}{Sections}
\Crefname{table}{Table}{Tables}
\crefname{table}{Tab.}{Tabs.}
\crefname{lstlisting}{Snippet}{Snippets}

\title{Leveraging Large Language Models for Multimodal Search} 

\author{
Oriol Barbany$^{1\dagger}$
\hspace{2ex} Michael Huang$^{2\ast}$
\hspace{2ex} Xinliang Zhu$^{2\ast}$
\hspace{2ex} Arnab Dhua$^{2}$ \\ \vspace{1ex}
\normalsize{${}^{1}$Institut de Robòtica i Informàtica Industrial, CSIC-UPC} \hskip3ex \normalsize{${}^{2}$Visual Search \& AR, Amazon}
}

\begin{document}

\input{figs/teaser}

\iftoggle{cvprfinal}{
\blfootnote{$\dagger$ Work performed during an internship at Amazon.}
\blfootnote{$^\ast$ Equal contribution.}
}{}

\glsresetall

\vspace{-1em}

\begin{abstract}
    Multimodal search has become increasingly important in providing users with a natural and effective way to express their search intentions. Images offer fine-grained details of the desired products, while text allows for easily incorporating search modifications.
    However, some existing multimodal search systems are unreliable and fail to address simple queries. The problem becomes harder with the large variability of natural language text queries, which may contain ambiguous, implicit, and irrelevant information. Addressing these issues may require systems with enhanced matching capabilities, reasoning abilities, and context-aware query parsing and rewriting.
    This paper introduces a novel multimodal search model that achieves a new performance milestone on the Fashion200K dataset \cite{fashion200k}. Additionally, we propose a novel search interface integrating \glspl{llm} to facilitate natural language interaction. This interface routes queries to search systems while conversationally engaging with users and considering previous searches.
    When coupled with our multimodal search model, it heralds a new era of shopping assistants capable of offering human-like interaction and enhancing the overall search experience.
\end{abstract}

\section{Introduction}
\label{sec:intro}

The \gls{cir} problem, also known as \gls{tgir}, involves finding images that closely match a reference image after applying text modifications. For instance, given a reference image of a blue dress and the instruction \texttt{"replace blue with red"}, the retrieved images should depict red dresses resembling the reference.

It is natural for users to search for products using information from multiple modalities, such as images and text. Enabling visual search allows for finding visually similar correspondences and obtaining fine-grained results. Otherwise, text-only search tools would require extensive textual descriptions to reach the same level of detail. Thus, it is more natural and convenient for users to upload a picture of their desired product or a similar version rather than articulating their search entirely in words.

Traditional search engines often struggle to deliver precise results to users due to the challenges posed by overly specific, broad, or irrelevant queries. Moreover, these engines typically lack support for understanding natural language text and reasoning about search queries while conversationally engaging with the user.

In the context of the Fashion200K benchmark \cite{fashion200k}, several existing approaches fail to retrieve the correct query among the top matches. Concretely, most of the baselines considered in this work fail to retrieve the correct image among the top 10 matches in 60\% of the cases, as shown in our results in \cref{sec:experiments_f200k}.

In this paper, we propose to leverage pretrained large-scale models that can digest image and text inputs. We focus on improving the performance on the Fashion200K dataset \cite{fashion200k} and achieve state-of-the-art results that improve upon previous work by a significant margin.
However, all the queries in Fashion200K follow the simple formatting \texttt{"replace \{original\_attribute\} with \{target\_attribute\}"}, which impedes generalizing to natural language text. For this reason, we develop a novel interactive multimodal search solution leveraging recent advances in \glspl{llm} and vision-language models that can understand complex text queries and route them to the correct search tool with the required formatting.
Leveraging \glspl{llm} facilitates digesting natural language queries and allows taking contextual information into account. Moreover, the length of the context recent \glspl{llm} can consider allows for incorporating information from previous interactions. We include a high-level overview of our approach in \cref{fig:first_figure}. The main contributions of this work include:

\begin{itemize}
    \item \textbf{Improved Multimodal Search: }We introduce a method that adapts foundational vision and language models for multimodal retrieval, which achieved state-of-the-art results on Fashion200k. We present the technical details in \cref{sec:search_model} and discuss the experimental results in \cref{sec:experiments_f200k}.
    \item \textbf{Conversational Interface: }We propose an interface that harnesses state-of-the-art \glspl{llm} to interpret natural language inputs and route formatted queries to the available search tools. We describe the details of the backend in \cref{sec:interface} and include examples in \cref{sec:interface_experiments}.
\end{itemize}

\section{Related work}
\label{sec:related_work}

When tackling the \gls{cir} problem, the TIRG model \cite{vo_composing_2019} computes an image representation and modifies it with a text representation on the same space rather than fusing both modalities to create a new feature as in most of the other works. Crucially, this method is trained first on image retrieval and gradually incorporates text modifications.

The VAL framework \cite{Chen_2020_CVPR} is based on computing image representations at various levels and using a transformer \cite{vaswani2017attention} conditioned on language semantics to extract features. Then, an objective function evaluates the feature similarities hierarchically.

The text and image encoders of a CLIP model \citep{clip} can be used for zero-shot retrieval with a simple \gls{mlp} \citep{pic2word} and leveraging \glspl{llm} \citep{searle}. Another approach is to perform a late fusion of CLIP embeddings \citep{baldrati_effective_2022}, which can be improved by fine-tuning the CLIP text encoder \citet{baldrati_conditioned_2022}. The hypothesis is that image and text embeddings obtained by CLIP are aligned, while the \gls{cir} problem requires a text representation that expresses differences \wrt the image representation.

CosMo \cite{CoSMo2021_CVPR} independently modulates the content and style of the reference image based on the modification text. This work assumes that style information is removed by simply performing instance normalization on the image features. With this assumption in mind, the normalized features are fed to the content modulator, which transforms them conditioned on text features. Then, the output of the content modulator is given to the style modulator, which along with the text features and the channel-wise statistics of the normalization, obtains the final representation.

FashionVLP \citep{goenka_fashionvlp_2022} is based on extracting image features using a pretrained feature extractor, not only on the whole image but also on the cropped clothing, fashion landmarks, and regions of interest. The obtained image representations are concatenated with object tags extracted with an object detector, a class token, and the word tokens computed using BERT \citep{devlin2018bert}.

\input{figs/full_architecture}

An alternative to tackle the problem of generic visual feature extractors not focusing on fashion-specific details without using the multiple inputs required in \citet{goenka_fashionvlp_2022}, is proposed in FashionSAP \citep{FashionSAP}. FashionSAP leverages the FashionGen dataset \citep{fashiongen} for fine-grained fashion vision-language pretraining. To do that, \citet{FashionSAP} use a multi-task objective composed of retrieval and language modeling losses. \gls{cir} is then solved by fusing text and image features using multiple cross-attention layers, and the tasks included in the training objective are solved using different heads for each task.

CompoDiff \citep{gu_compodiff_2023} proposes to solve the \gls{cir} using a denoising transformer that provides the retrieval embedding conditioned on features of the reference image and the modifying text. Similarly to \citet{stable_diffusion}, the diffusion process is performed in the latent space instead of the pixel space. Given that CompoDiff is a data-hungry method, \citet{gu_compodiff_2023} create a synthetic dataset of 18 million image triplets using StableDiffusion \citep{stable_diffusion} for its training. CompoDiff performs better when using text features obtained with a T5-XL model \citep{t5} in addition to the text representations obtained with \citep{clip}.

\citet{fromage} uses a frozen \gls{llm} to process the input text and visual features that have been transformed with a learned linear mapping as in LLaVA \citep{llava}. To counteract the inferior expressiveness of causal attention over its bidirectional counterpart, \citet{fromage} append a special \texttt{[RET]} token at the end of the outputs that allows the \gls{llm} to perform an extra attention step over all tokens. The hidden representations of \texttt{[RET]} are then mapped to an embedding space that is used for retrieval.

\citet{simat} tackle a similar problem in which the transformation query is not a single word but a tuple of two words corresponding to the original and target attributes. As an example, for a reference image with caption \texttt{"A cat is sitting on the grass"}, a source text \texttt{"cat"} and a target text \texttt{"dog"}, the model should be able to retrieve images of dogs sitting on the grass.

\section{Method}
\label{sec:method}

In this section, we propose a model to perform an image search merging text and image inputs in \cref{sec:search_model}. While this model outperforms alternative approaches by a large margin, it is trained on a dataset with specific formatting (see \cref{sec:experiments_f200k}). Instead of artificially augmenting the vocabulary seen during training as in \citet{gu_compodiff_2023}, we propose a conversational interface orchestrated by a \gls{llm} that can structure the queries to a format understandable to our multimodal search model.

\cref{sec:interface} describes the principles of our approach. The proposed framework offers a modular architecture that allows interchanging search models with different formatting constraints while providing enhanced natural language understanding, a working memory, and a human-like shopping assistant experience.

\subsection{Improved multimodal search}
\label{sec:search_model}

In the \gls{cir} problem, a dataset $\cD$ is composed of triplets with reference and target image as well as a modifying text, \ie, $\cD:=\{(\xx_{\text{ref}}^{(i)}, \xx_{\text{trg}}^{(i)}, \ttt^{(i)})\}_{i\in[n]}$. The objective is to learn the transformations
\begin{align}
    \cF: \xx_{\text{ref}} \times \ttt \mapsto \Psi \qquad ; \qquad \cG: \xx_{\text{trg}} \mapsto \Psi
\end{align}
along with a metric space $(\Psi, d)$ with fixed $d:\Psi\times \Psi \to \R$ such that
\begin{align}
    d(\cF(\xx_{\text{ref}}, \ttt), \cG(\xx_{\text{trg}})) < d(\cF(\xx_{\text{ref}}, \ttt), \cG(\xx_{\text{trg}}'))
\end{align}
if $\xx_{\text{ref}}$ after applying the modifications described by $\ttt$ is semantically more similar to $\xx_{\text{trg}}$ than it is to $
\xx_{\text{trg}}'$ \cite{metric_learning_book}. Commonly to other works \cite{triplet_loss,margin,divide_and_conquer}, we normalize the space $\Psi$to the unit hypersphere for training stability, and choose $d$ to be the cosine distance.

In this work, we use off-the-shelf foundational models for vision and language to compute the transformation $\cF$. Concretely, we use an architecture similar to BLIP2 \citep{blip2} and adapt it for the \gls{cir} problem. BLIP2 \cite{blip2} uses a module referred to as the \gls{qformer}, which allows ingesting image features obtained by a powerful feature extractor. These image features provide fine-grained descriptions of the input product and are transformed into the space of text embeddings of a \gls{llm}. Then, the \gls{llm} processes the fused text and image embeddings. 

The \gls{qformer} consists of two transformer submodules sharing the same self-attention layers to extract information from the input text and the image features. The image transformers also contain a set of learnable query embeddings, which can be interpreted as a form of prefix tuning \citep{li_prefix-tuning_2021}.

To generate image-only search embeddings using our model, one simply needs to input the images into the model and provide an empty string as the input text. Intuitively, this processes the images without any text modifications. In other words, we use
\begin{align}
    \cG(\xx) := \cF(\xx, \texttt{""})
\end{align}

We illustrate the proposed architecture for $\cF$ in \cref{fig:full_architecture}. We use the image part of the CLIP \citep{clip} model to obtain visual features and a T5 model \citep{t5} as \gls{llm} to process the modifying text and the visual features processed by the \gls{qformer}.

We initialize the model using the BLIP2 weights with all the parameters frozen. The pretrained weights perform the task of image captioning, which is different from the task we are trying to solve. Instead, we define a new task that we refer to as \textit{composed captioning}. The objective of this task is to generate the caption of the product that we would obtain by merging the information of the product in the input image and the text modifications.

We hypothesize that if the proposed model can solve the problem of \textit{composed captioning}, the information captured by the \gls{llm} is enough to describe the target product. Intuitively, similarity search happens at a latent space close to the final text representations, making the \gls{cir} problem closer to the task of text-to-text retrieval. However, as the proposed model is able to capture fine-grained information by leveraging powerful visual representations, we are able to obtain an impressive retrieval performance. This is expected as the BLIP2 achieves state-of-the-art performance on \gls{vqa} benchmarks, showing that image information can be effectively captured.

To adapt the \gls{llm} to this task while retaining its knowledge, we applied \gls{lora} \citep{hu_lora_2021}  to the query and value matrices of all the self-attention and cross-attention layers of the \gls{llm}. \gls{lora} \cite{hu_lora_2021} learns a residual representation on top of some layers using matrices with low rank. Theoretically, this is supported by the fact that \glspl{llm} adapted to a specific task have low intrinsic dimension \citep{aghajanyan-etal-2021-intrinsic}, and in practice it allows training with low computational resources and limited data. Moreover, only modifying a few parameters reduces the risk of catastrophic forgetting, observed in some studies where full fine-tuning of an \gls{llm} decreases the performance compared to using it frozen or fine-tuning it with parameter-efficient techniques \citep{fromage,ia3}.

The hidden states of the T5 decoder are a sequence of tensors. Instead of using a class-like token as in \citet{fromage} to summarize the information along the temporal dimension, we perform an average followed by layer normalization \citep{ba2016layer}. This technique was utilized in EVA \citep{fang2023eva}, which improves over CLIP \citep{clip} in several downstream tasks. The result is then projected to the embedding dimension using a ReLU-activated \gls{mlp} and followed by normalization.

We train the model using a multi-task objective involving the InfoNCE loss \citep{infonce}, a lower bound on the mutual information \citep{ALBEF}, as retrieval term:
\begin{align}
    \begin{split}
    \cL_{\text{InfoNCE}} := \E_i \left[ \log\frac{\exp\left(S_{i, i} \cdot \tau \right)  }{\sum_{j}\exp\left(S_{i, j}\cdot \tau \right)} \right] \\
    S_{i, j} := \lin{\cF(\xx_{\text{ref}}^{(i)}, \ttt^{(i)}), \cG(\xx_{\text{trg}}^{(j)})}\,,
    \end{split}
    \label{eq:infonce}
\end{align}
where $\tau$ is a learnable scaling parameter. Practically, given that our model has many parameters, the maximum batch sizes we can achieve have an order of magnitude of hundreds of samples. Given that this can affect the retrieval performance due to a lack of negative samples, we maintain a cross-batch memory as proposed in \citet{wang2020xbm} and use it for the computation of \cref{eq:infonce}.

On top of that, we add a standard maximum likelihood as a language modeling term $\cL_{\text{LM}}$. We compute this objective using teacher forcing \citep{teacher_forcing}, based on providing the ground-truth outputs of previous tokens to estimate the next token, and cross-entropy loss. The final loss is
\begin{align}
    \cL =\cL_{\text{LM}}  + \omega \cL_{\text{InfoNCE}}\,,
\end{align}
where $\omega$ is a hyperparameter determining the relative importance of the retrieval task.

\subsection{Conversational interface}
\label{sec:interface}

Inspired by Visual ChatGPT \citep{visualchatgpt}, we connect a user chat to a prompt manager that acts as a middle-man to a \gls{llm} and provides it with access to tools. Differently from \citet{visualchatgpt}, these tools are not only to understand and modify images but also to perform searches with both unimodal and  multimodal inputs.

From the user's perspective, the proposed framework allows implicitly using a search tool without requiring any input pattern. For example, interacting with a model like SIMAT \cite{simat} could be unintuitive as it requires two words with the original and target attributes. We trained our multimodal search model on Fashion200K \cite{fashion200k}, which only contains inputs of the form \texttt{"replace \{original\_attribute\} with \{target\_attribute\}"} (see \cref{sec:experiments_f200k}). We could formulate this prompt using the same inputs that a model like SIMAT requires and thus modify them to match the training distribution of our model.

Since the \glspl{llm} can only ingest text information, we add image understanding tools to provide information about the images and their content, as well as search tools:

\par\textbf{Image search: }Image-only search based on CLIP \cite{clip} image embeddings. We use this tool internally when a user uploads an image to show an initial result to users, which may inspire them to write the follow-up queries. The descriptions of the search results are provided to the \gls{llm} to enable \gls{rag} \cite{rag}

\par\textbf{Multimodal search: }The input of the multimodal search tool is an image and two text strings expressing the original and target attributes. We use our model and feed it the Fashion200K \cite{fashion200k} prompt created from these attributes. 

\par\textbf{\gls{vqa} model: }We use the BLIP \cite{blip} pretrained base model\footnote{\url{https://huggingface.co/Salesforce/blip-vqa-base}} to facilitate image understanding to the \gls{llm}.

Our approach to providing image information to the \gls{llm} is similar to LENS \citep{lens}, as it is a training-free method applicable to any off-the-shelf \gls{llm}.

\subsubsection{Workflow}

In this section, we describe the main events in the interface and the triggered actions.

\par \textbf{Start: }When a new user starts a new session, we create a unique identifier used to set a dedicated folder to store images and initialize the memory to store the context. The memory contains a conversation where the lines prefixed with \texttt{"Human:"} come from the user, and those starting with \texttt{"AI:"} are outputs of the \gls{llm} shown to the user.

\par \textbf{Image input: }When a user uploads an image, we store it in the session folder using file names with sequential numerical identifiers, \ie, \texttt{IMG\_001.png}, \texttt{IMG\_002.png}, \texttt{IMG\_003.png}, \etc. Then we add a fake conversation to the memory:
\begin{lstlisting}
Human: I provided a figure named {image_filename}. {description}
AI: Provide more details if you are not satisfied with the results.
\end{lstlisting}
where \texttt{description} is the text output of the search action.

\par \textbf{Search: }Every time a search tool is used, the results are shown to the user in a carousel of images. Additionally, we add the following information to the memory that will be provided to the \gls{llm} once invoked
\begin{lstlisting}
Top-{len(image_descriptions)} results are: {image_descriptions}.
\end{lstlisting}
which contains the descriptions of the top retrieved images. These details help the \gls{llm} understand the fine-grained details (\eg, brand, product type, technical specifications, color, \etc) and the multimodal search intention. We can interpret this as a form of \gls{rag} \citep{rag}. \gls{rag} is based on using an external knowledge base for retrieving facts to ground \glspl{llm} on the most accurate and up-to-date information.

\par \textbf{Text input: }Every time the user provides some text input, we invoke the \gls{llm}through the prompt manager. In this stage, the \gls{llm} can communicate directly to the user or use special formatting to call some tools. If the \gls{llm} wants to perform a multimodal search, it can typically find the target attribute in the text input, which only needs to be formatted and simplified. However, in most cases, the original attribute is not included in the input text as it is implicit in the image. Generally, the descriptions contain enough information to perform the query. Otherwise, the \gls{llm} can use the \gls{vqa} model to ask specific questions about the image.

\subsubsection{Prompt manager}
\label{sec:prompt_manager}

The prompt manager implements the workflow described in the previous section and empowers the \gls{llm} with access to different tools. The tool calls are coordinated by defining a syntax that processes the output of the \gls{llm} and parses the actions and the text visible to the user in the chat.

Every time the \gls{llm} is triggered, the prompt manager does so with a prompt that includes a description of the task, formatting instructions, previous interactions, and outputs of the tools.

We crafted a task description that specifies that the \gls{llm} can ask follow-up questions to the customers if the search intents are unclear or the query is too broad. In the prompt, we also include examples of use cases written in natural language. The formatting instructions describe when the \gls{llm} should use a tool, which are the inputs, how to obtain them, and what are the tool outputs.

For each tool, we have to define a name and a description that may include examples, input and output requirements, or cases where the tool should be used.

In this work, we test two prompt managers:

\par\textbf{Langchain \citep{langchain}: }We take the Langchain prompts from Visual ChatGPT \citep{visualchatgpt} and adapt them to our task. The syntax to use a tool is:

\begin{lstlisting}
Thought: Do I need to use a tool? Yes
Action: Multimodal search
Action Input: IMG_001.png;natural;black
\end{lstlisting}

\par\textbf{Our prompt manager: }Inspired by the recent success of visual programming \citep{visprog,vipergpt}, we propose to use a syntax similar to calling a function in programming languages:

\begin{lstlisting}
SEARCH(IMG_001.png;natural;black)
\end{lstlisting}

In \cref{fig:first_figure}, we illustrate an example of a conversation and the actions that the prompt manager and the \gls{llm} trigger.

Visual programming typically performs a single call to a \gls{llm}, and the output is a single action or a series of actions whose inputs and outputs can be variables defined on the fly by other functions. While Langchain \citep{langchain} allows performing multiple actions, it requires executing them one at a time. When the \gls{llm} expresses the intention to use a tool, Langchain calls the tool and prompts the \gls{llm} again with the output of such a tool. The visual programming approach only invokes the \gls{llm} once, saving latency and possible costs attributed to API calls. However, in visual programming, the \gls{llm} cannot process the output of tools but only use their outputs blindly. For the sake of simplicity, we restrict the custom prompt manager to handle single actions, but this could easily be extended following \citet{visprog,vipergpt}.

Additionally, we propose to include \gls{cot} \citep{cot1,cot2,cot3,multimodal_cot}. \gls{cot} is a technique that enforces that the \gls{llm} reasons about the actions that should be taken. This simple technique has reportedly found numerous benefits. Following the example above, the complete output expected by the \gls{llm} would be as follows:

\begin{lstlisting}
Thought: I can see that human uploaded an image of a deep v-neck tee. From the results, the color of the tee is natural. The user wants the color to be black instead. I have to call search.
Action: SEARCH(IMG_001.png;natural;black)
\end{lstlisting}

While Langchain and our prompt manager use the special prefix \texttt{"Thought"} to handle certain parts of the query, their purposes are distinct. In Langchain, the prefix is used to parse lines in the \gls{llm} output. If a line starts with this prefix, Langchain expects to find the question \texttt{"Do I need to use a tool?"} followed by \texttt{"Yes"} or \texttt{"No"}, indicating whether a tool should be used. In contrast, our novel prompt manager does not impose any specific format on lines starting with the \texttt{"Thought"} prefix. Instead, these lines are solely dedicated to incorporating \gls{cot} reasoning.

\section{Experiments}
\label{sec:experiments}

\subsection{Multimodal search on Fashion200K}
\label{sec:experiments_f200k}

\input{tabs/fashion200k}

\input{figs/fashion200k_qualitative}

\textbf{Implementation details: }We use the Flan T5 XL model \cite{flan_t5}, which is a 3 billion parameter \gls{llm} from the T5 family \cite{t5}, finetuned using instruction tuning \cite{flan}. We obtain the visual features with CLIP-L model \cite{clip}, a model with patch size 14 and 428 million parameter. In total, the model has around 3.5 million parameter, which requires splitting the model across different GPUs for training. Concretely, we use 8 NVIDIA V100 GPUs.

\gls{lora} is performed with a rank of $r=16$, scaling $\alpha=32$ and dropout of 0.5 on the query and value matrices of the attention layers of the \gls{llm}. The hidden representation obtained from the \gls{llm} is transformed with a linear layer of size 1024, passed through a ReLU activation, and then transformed with another linear layer that yields an embedding of size 768. Such an embedding is normalized to have unit norm and used for retrieval.

We optimize the model with AdamW \cite{adamw} with a learning rate of $10^{-5}$ and weight decay of 0.5 for a total of 300 epochs. The learning rate is linearly increased from 0 to the initial learning rate during the first 1000 steps. We set the weight of the language modeling loss as $\omega=1$. The effective batch size considering all the GPUs is 4,096, and the total number of embeddings included in the cross-batch memory \citet{wang2020xbm} is 65,536.

\textbf{Dataset: }The Fashion200K \cite{fashion200k} is a large-scale fashion dataset crawled from online shopping websites. The dataset contains over 200,000 images with paired product descriptions and attributes. All descriptions are fashion-specific and have more than four words, \eg, ``Beige v-neck bell-sleeve top". Similarly to \citet{vo_composing_2019}, text queries for the \gls{cir} problem are generated by comparing the attributes of different images and finding pairs with one attribute difference. Then, a query is formed as
\texttt{"replace \{original\_attribute\} with \{target\_attribute\}"}.

When trained on Fashion-200K \citep{fashion200k}, our method achieves state-of-the-art results, improving the retrieval performance of competitive methods by 20\% recall at positions 10 and 50. \cref{tab:fashion200k} includes the comparison with some of the \gls{cir} methods reviewed in \cref{sec:related_work} \cite{vo_composing_2019,Chen_2020_CVPR,CoSMo2021_CVPR,goenka_fashionvlp_2022}, as well as the visual reasoning-based baselines RN \cite{rn_fashion200k}, MRN \cite{mrn}, and FiLM \cite{perez2018film}.

One of the reasons is that the model can exploit the image and text understanding prior of a foundational model that can perform image captioning, and adapt it for the related task of \textit{composed captioning}. The hidden representations of the model contain enough information to describe the target image and are effectively used for that purpose. Adapting to this new task becomes easier given the specific formatting of the modifying text, which facilitates extracting the important parts of the query.

The results show that it is possible to distill knowledge from a large vision and language model trained on large-scale datasets. While our model has billions of parameters, which is far more than the other models, we are able to learn a new task similar to the ones that the pretrained model could solve with only learning a few parameters consisting of a very small percentage of the total model size.

We include some qualitative examples in \cref{fig:fashion200k_qualitative}. These show that our model can successfully incorporate text information and modify the internal description formed about the input image. The successful results in \cref{fig:success} show that the proposed model retrieves visually similar and can incorporate modifications of different attributes such as the color and the material.

The failures in \cref{fig:failure} show that all the first retrieve results satisfy the search criteria, with some of them even belonging the same product. This hints at our model having an even better performance in practice than what the benchmark reflects.

Overall, we can see from all the qualitative examples that all the top-ranked results are relevant. The only exception is the inclusion of the reference image, which is a common error in retrieval systems given that the search embedding is computed from such an image.

\input{figs/interface}

\subsection{Search interface}
\label{sec:interface_experiments}

One of the key drivers of performance is based on reformulating the examples. While the examples in Langchain are written using natural language, we advocate for using \gls{llm} model instructions. In this sense, the examples contain exactly the input that the \gls{llm} would receive including the product type, top-$k$ product titles, and user input. Such examples also contain the expected model output including the \gls{cot} reasoning and the action itself. This reinforces the format instructions and the benefits of \gls{rag}.

Note that the proposed reformulation introduces some redundancy \wrt the Langchain formatting instructions. Additionally, it requires to allocate much more space for examples. Despite these considerations, we find our approach beneficial. For a fair comparison, we also limit the full prompt to fit the context of the smallest \gls{llm} and empirically find that allocating more space to examples is beneficial even if this is at the cost of removing the prefix.

We tested different \glspl{llm} for the search interface. Among all, GPT-3 \citep{gpt3}, concretely the \texttt{text-davinci-003} model, was empirically found to be the best performing. \cref{fig:interface} shows an example of our conversational interface displaying a real example in which composed retrieval is performed.

Besides GPT-3 \cite{gpt3}, we compared different open-source models from the \texttt{transformers} library \citep{wolf-etal-2020-transformers}. Surprisingly, these models performed poorly. Digging into the outputs of the \glspl{llm} we could see that one of the failure cases of Fastchat \citep{fastchat} had the following output:

\begin{lstlisting}
Thought: Do I need to use a tool? Yes
Action: Multimodal altering search
Action Input: image_file: IMG_001.png, attribute value of the product in the image: chair, desired attribute value: sofa
\end{lstlisting}

While the former contains the correct action to take and the correct inputs, \ie image file, negative text and positive text, it is not correctly formatted for Langchain. Instead, GPT-3 \citep{gpt3} is able to generate a correctly formatted output:

\begin{lstlisting}
Thought: Do I need to use a tool? Yes
Action: Multimodal altering search
Action Input: IMG_001.png;chair;sofa
\end{lstlisting}

This example shows that FastChat \cite{fastchat} has the knowledge to perform a successful query but struggles to use the complicated formatting of Langchain. This example is the main motivation why we developed the novel prompt manager in \cref{sec:prompt_manager}.

\section{Limitations}
\label{sec:limiations}

The model in \cref{sec:search_model} can achieve an impressive performance on Fashion200K. As discussed in \cref{sec:experiments_f200k}, the characteristics of this dataset are ideal for our model to excel but may hinder generalizing to natural language queries. This is solved with our conversational interface, but the current setup is restricted to modifying a single attribute at a time.

Using hard prompts to encode the task description is simple and applicable to black-box models such as \glspl{llm} accessed through an API. However, it reduces the effective context length of \glspl{llm} and requires prompt engineering, which is a tedious process.

 Although \glspl{llm} have a large context size, the prompt yields an effective input size that is relatively small, and the memory rapidly fills up. In practice, the memory gets truncated if conversations are too long, hence discarding the first interactions.

\section{Conclusions}

This paper presents a comprehensive pipeline to perform image retrieval with text modifications, addressing the \gls{cir} problem. Our novel composed retrieval model, built upon the BLIP2 architecture \citep{blip} and leveraging \glspl{llm}, has demonstrated superior performance on the Fashion200K dataset \cite{fashion200k} compared to previous models.

In this work, we also describe the integration of \glspl{llm} into a search interface, offering a conversational search assistant experience that enhances user interaction. We implement a prompt manager to enable using small \glspl{llm} and incorporate the \gls{cot} \cite{cot1,cot2} and \gls{rag} \cite{rag} techniques to improve system performance.

Our experiments underscore the importance of addressing inherent challenges in multimodal search, including enhancing matching capabilities and handling ambiguous natural language queries.

\iftoggle{cvprfinal}{
\section*{Acknowledgments}
The authors acknowledge René Vidal for constructive discussions. O.B. is part of project SGR 00514, supported by Departament de Recerca i Universitats de la Generalitat de Catalunya.
}{}

{
    \small
    \bibliographystyle{ieeenat_fullname}
    \bibliography{references}
}

\end{document}

%% file: preamble.tex
\usepackage[table,dvipsnames]{xcolor}

\newcommand\blfootnote[1]{%
  \begingroup
  \renewcommand\thefootnote{}\footnote{#1}%
  \addtocounter{footnote}{-1}%
  \endgroup
}

\usepackage{listings}
\definecolor{codegreen}{rgb}{0,0.6,0}
\definecolor{codegray}{rgb}{0.5,0.5,0.5}
\definecolor{codepurple}{rgb}{0.58,0,0.82}
\definecolor{backcolour}{rgb}{0.95,0.95,0.92}
\lstdefinestyle{mystyle}{
    backgroundcolor=\color{backcolour},   
    commentstyle=\color{codegreen},
    keywordstyle=\color{magenta},
    stringstyle=\color{codepurple},
    basicstyle=\ttfamily\scriptsize,
    breakatwhitespace=false,         
    breaklines=true,                 
    captionpos=b,                    
    keepspaces=true,                 
    showspaces=false,                
    showstringspaces=false,
    showtabs=false,                  
    tabsize=2,
    showlines=true
}
\usepackage[acronym]{glossaries}
\newacronym{llm}{LLM}{Large Language Model}
\newacronym{vqa}{VQA}{Visual Question Answering}
\newacronym{rag}{RAG}{Retrieval Augmented Generation}
\newacronym{mlp}{MLP}{Multi-Layer Perceptron}
\newacronym{relu}{ReLU}{Rectified Linear Unit}
\newacronym{peft}{PEFT}{Parameter-Efficient Fine-Tuning}
\newacronym{lora}{LoRA}{Low-Rank Adaptation}
\newacronym{cot}{COT}{Chain-of-Thought}
\newacronym{qformer}{Q-Former}{Querying transFormer}
\newacronym{vit}{ViT}{Vision Transformer}
\newacronym{cir}{CIR}{Composed Image Retrieval}
\newacronym{tgir}{TGIR}{Text-Guided Image Retrieval}

\usepackage{booktabs}       
\usepackage{amsfonts}       
\usepackage{graphicx}
\usepackage{amsmath}
\usepackage{amssymb}

\providecommand{\R}{\mathbb{R}} 

\providecommand{\lin}[1]{\ensuremath{\left\langle #1 \right\rangle}}

\DeclareMathOperator{\E}{{\mathbb E}}

\providecommand{\ttt}{\mathbf{t}}

\providecommand{\xx}{\mathbf{x}}

\providecommand{\cD}{\mathcal{D}}

\providecommand{\cF}{\mathcal{F}}
\providecommand{\cG}{\mathcal{G}}

\providecommand{\cL}{\mathcal{L}}


%% file: figs/teaser.tex
\twocolumn[{%
\renewcommand\twocolumn[1][]{#1} %
\maketitle
\begin{center}
    \centering
    \vspace{-2em}
    \captionsetup{type=figure}
    \includegraphics[width=\textwidth]{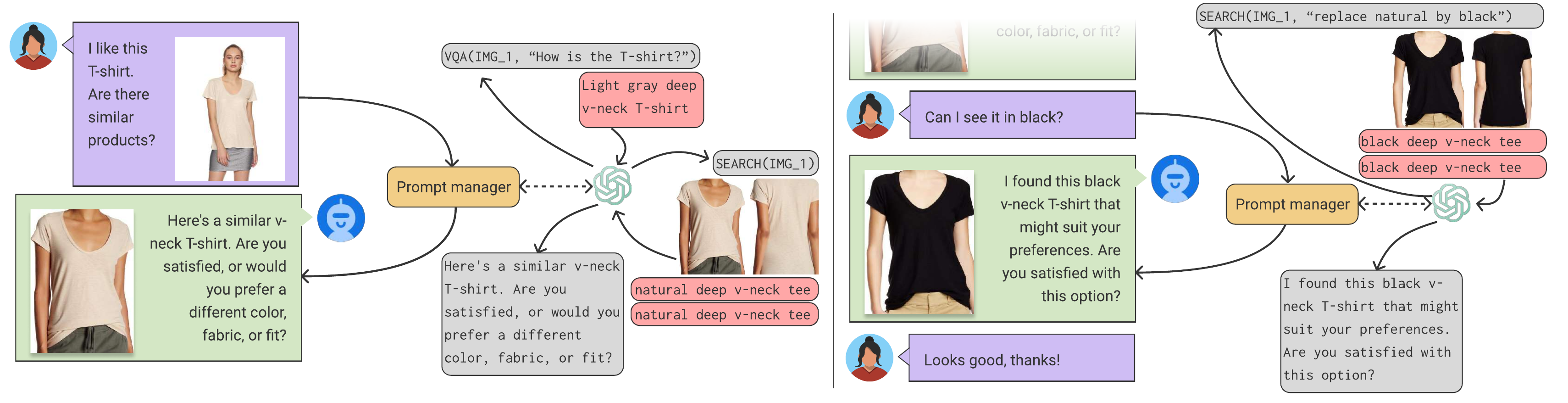}
    \captionof{figure}{\textbf{Overview}: This paper introduces a comprehensive pipeline for multimodal search, presenting a novel composed retrieval model that outperforms previous approaches significantly. Additionally, we propose a system that utilizes a \gls{llm} as an orchestrator to invoke both our proposed model and other off-the-shelf models. The resulting search interface offers a conversational search assistant experience, integrating information from previous queries and leveraging our novel model to enhance search capabilities.}
    \label{fig:first_figure}
\end{center}
}]

%% file: figs/full_architecture.tex
\begin{figure*}[t]
    \centering
    \includegraphics[width=\textwidth]{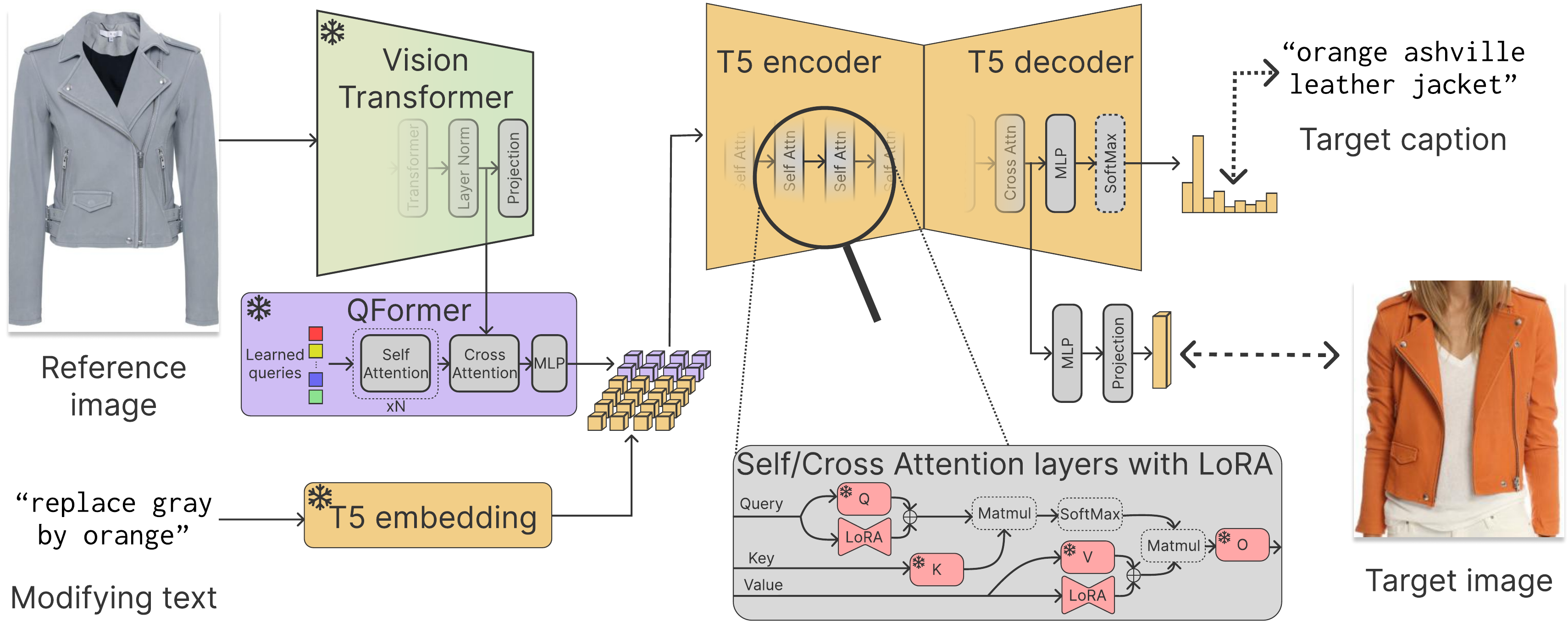}
    \vspace{-2em}
    \caption{\textbf{Proposed architecture}: We extract visual features from the reference image $\xx_{\text{ref}}$ using a Vision Transformer \cite{dosovitskiy2020vit}, specifically, a pretrained CLIP \citep{clip} model with frozen weights. We extract features before the projection layer, which are then processed using a \gls{qformer}, which performs cross-attention with a set of learned queries. The resulting output of the \gls{qformer} is concatenated with the embeddings obtained from the modifying text ($\ttt$), which expresses a modification in the reference image. Subsequently, all this information is fed into a T5 model \cite{t5}, an encoder-decoder \gls{llm}. We employ \gls{lora} \cite{hu_lora_2021} to  learn low-rank updates for the query and value matrices in all attention layers, while keeping the rest of the parameters frozen. The output of the \gls{llm} yields a probability distribution from which a sentence is generated. To ensure alignment with the target caption (\ie, the caption of the target image $\xx_{\text{trg}}$, which corresponds to the caption of the reference image after incorporating the text modifications), a language modeling loss is used. The hidden states of the \gls{llm} are then projected into a space of embeddings used for retrieval. A retrieval loss term pushes together the embedding of the target image $\cG(\xx_{\text{trg}})$ and that obtained using the reference image and the modifying text $\cF(\xx_{\text{ref}}, \ttt)$.}
    \label{fig:full_architecture}
    \vspace{-1em}
\end{figure*}

%% file: tabs/fashion200k.tex
\begin{table}[t]
    \centering
    \caption{\textbf{Quantitative results: }Recall@$k$ on the Fashion200K dataset \citep{fashion200k}. Our method is able to successfully fuse image and text information and generate a representation that is useful to caption the resulting image and generate an embedding for retrieval purposes. Best results shown in \textbf{boldface}.}
    \label{tab:fashion200k}
    \vspace{-1em}
    \begin{tabular}{lccc}
        \toprule
         Method $\downarrow$ & R@10 & R@50 & Average \\
         \midrule
         RN \citep{rn_fashion200k} & 40.5 & 62.4&  51.4 \\
         MRN \citep{mrn} &40.0& 61.9 &50.9 \\
         FiLM \citep{perez2018film} & 39.5& 61.9& 50.7 \\
         TIRG \citep{vo_composing_2019} &42.5& 63.8 &53.2 \\
         CosMo \citep{CoSMo2021_CVPR} & 50.4 & 69.3 &59.8 \\ 
         FashionVLP \citep{goenka_fashionvlp_2022} & 49.9 & 70.5 & 60.2 \\
         VAL \citep{Chen_2020_CVPR} & 53.8 &73.3& 63.6 \\
         \midrule
         Ours & \textbf{71.4} & \textbf{91.6} & \textbf{81.5} \\
        \bottomrule
    \end{tabular}
    \vspace{-1em}
\end{table}

%% file: figs/fashion200k_qualitative.tex
\begin{figure*}[t]
    \centering
    \begin{subfigure}{.48\textwidth}
        \centering
        \includegraphics[width=\textwidth]{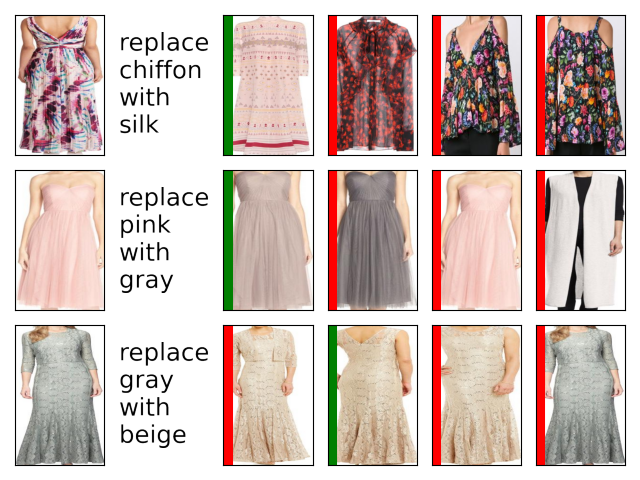}
        \caption{Successful examples}
        \label{fig:success}
    \end{subfigure}
    \hfill
    \begin{subfigure}{.48\textwidth}
        \centering
        \includegraphics[width=\textwidth]{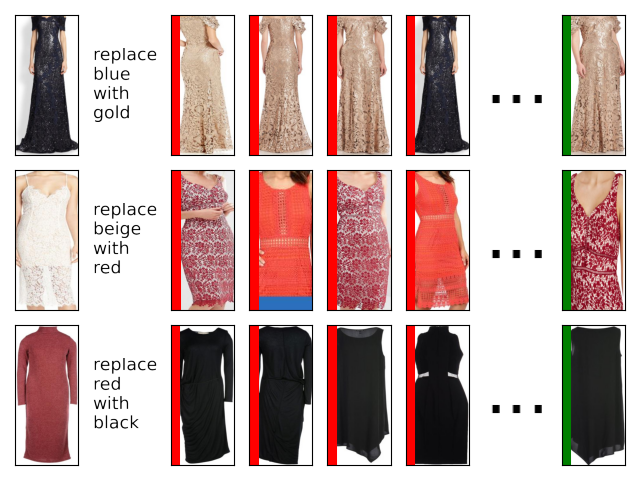}
        \caption{Failure examples}
        \label{fig:failure}
    \end{subfigure}
    \caption{\textbf{Qualitative results}: Examples of queries of the Fashion-200k dataset \citep{fashion200k} and the 4 best matches. The correct matches are shown in {\color{ForestGreen}\textbf{green}} and incorrect ones in {\color{red}\textbf{red}}. In the succesful examples, we can see that our proposal is able to incorporate modifications to the input product involving changes to color and material among others. Despite not retrieving the correct products in the failure examples, almost all the retrieved images satisfy the search criteria.}
    \vspace{-1.5em}
    \label{fig:fashion200k_qualitative}
\end{figure*}

%% file: figs/interface.tex
\begin{figure*}
    \centering
    \includegraphics[width=\textwidth]{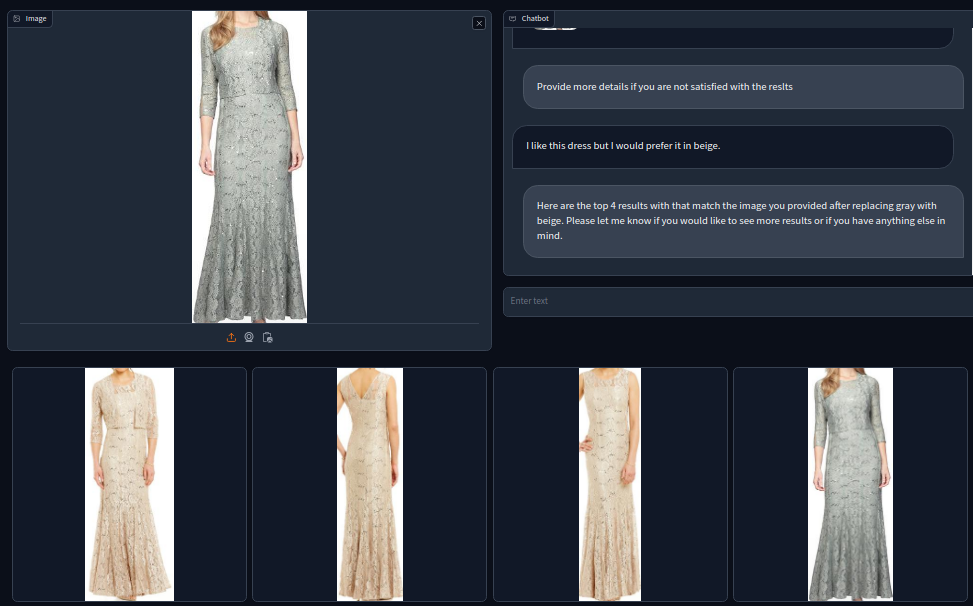}
    \vspace{-1em}
    \captionof{figure}{\textbf{Proposed conversational multimodal search system}: In this example, the user uploads an image from the Fashion200K dataset \cite{fashion200k} and provides text input intending to search an a dress similar to the product in the image but in a different color. An \gls{llm}, specifically GPT-3 \cite{gpt3}, processes the user's prompt and invokes our novel multimodal search model with the uploaded image and a formatted text query. The desired attribute indicated by the user is ``beige", which can be inferred from the text input. The original attribute is required by the prompt used during the training of our model and is correctly identified by the \gls{llm} as ``gray". In this case, the \gls{llm} can obtain this information leveraging the \gls{rag} based on obtaining the product descriptions of the first matches using image search with the uploaded picture. The conversational nature of the interactions with the user offers an improved search experience.}
    \vspace{-1em}
    \label{fig:interface}
\end{figure*}